\documentclass[sigconf]{acmart}

\AtBeginDocument{%
  \providecommand\BibTeX{{%
    \normalfont B\kern-0.5em{\scshape i\kern-0.25em b}\kern-0.8em\TeX}}}

\usepackage{xcolor}
\usepackage{booktabs}
\usepackage{epsfig}
\usepackage{graphicx}
\usepackage{amsmath}

\usepackage{amssymb}

\usepackage{verbatim}

\usepackage{url}
\usepackage{tabularx}
\usepackage{multirow}
\usepackage{subfigure}
\usepackage{amsmath,dsfont}
\usepackage{array}
\usepackage{enumitem}
\usepackage{pifont}
\usepackage{threeparttable}
\usepackage{balance}

\usepackage{verbatim}

\usepackage{colortbl}
\definecolor{mygray}{gray}{.75}

\usepackage{xcolor}
\usepackage{tikz-dependency}
\usepackage{pgfplots}
\usepackage{pgfplotstable}
\usepackage{makecell}

\newcommand{\black}[1]{\textcolor{black}{#1}}

\usepackage[normalem]{ulem}

\newcommand{\eg}{\emph{e.g.}~}
\newcommand{\etal}{\emph{et al.}~}
\newcommand{\ie}{\emph{i.e.,}~}

\copyrightyear{2022}
\acmYear{2022}
\setcopyright{acmcopyright}\acmConference[MM '22]{Proceedings of the 30th ACM International Conference on Multimedia}{October 10--14, 2022}{Lisboa, Portugal}
\acmBooktitle{Proceedings of the 30th ACM International Conference on Multimedia (MM '22), October 10--14, 2022, Lisboa, Portugal}
\acmPrice{15.00}
\acmDOI{10.1145/3503161.3548003}
\acmISBN{978-1-4503-9203-7/22/10}

\begin{document}
\title{Cross-Lingual Cross-Modal Retrieval with Noise-Robust Learning}


%



\author{Yabing Wang}
\affiliation{
  \institution{Zhejiang Gongshang University}
  \country{}
}

\orcid{0000-0001-7231-1260}
\authornote{Co-first authors.}

\author{Jianfeng Dong}
\authornotemark[1]
\affiliation{
  \institution{Zhejiang Gongshang University}
  \country{}
}
\orcid{0000-0001-5244-3274}

\author{Tianxiang Liang}
\affiliation{
  \institution{Zhejiang Gongshang University}
  \country{}
}

\author{Minsong Zhang}
\affiliation{
  \institution{Zhejiang Gongshang University}
  \country{}
}

\author{Rui Cai}
\authornote{Corresponding author (cairuics@gmail.com).}
\affiliation{
  \institution{Zhejiang Gongshang University}
  \country{}
}

\author{Xun Wang}
\affiliation{
  \institution{Zhejiang Gongshang University}
  \country{}
}

\renewcommand{\shortauthors}{Yabing Wang et al.}

\begin{abstract}
Despite the recent developments in the field of cross-modal retrieval, there has been less research focusing on low-resource languages due to the lack of manually annotated datasets.
In this paper, we propose a  noise-robust cross-lingual  cross-modal retrieval method for low-resource languages.
To this end, we use Machine Translation (MT)
to construct pseudo-parallel sentence pairs for low-resource languages.
However, as MT is not perfect, it tends to introduce noise during translation, rendering textual embeddings corrupted 
and thereby compromising the retrieval performance.
To alleviate this, we introduce a multi-view
self-distillation method to learn noise-robust target-language representations, which employs a cross-attention module to generate soft pseudo-targets to provide direct supervision from the
similarity-based view and feature-based view.
Besides, inspired by the back-translation in unsupervised MT, we minimize the semantic discrepancies
between origin sentences and back-translated sentences to further improve the noise robustness of the textual encoder.
Extensive experiments are conducted on three video-text and image-text cross-modal retrieval benchmarks across different languages, and the results demonstrate that our method significantly improves the overall performance without using extra human-labeled data.
In addition, equipped with a pre-trained visual encoder from a recent vision-and-language pre-training framework, \ie CLIP, our model achieves a significant performance gain,  showing that our method is compatible with popular pre-training models. Code and data are available at \url{https://github.com/HuiGuanLab/nrccr}.

\end{abstract}




\maketitle

\section{Introduction}\label{sec:introduction}

\begin{figure}[tb!]
\centering\includegraphics[width=0.95\columnwidth]{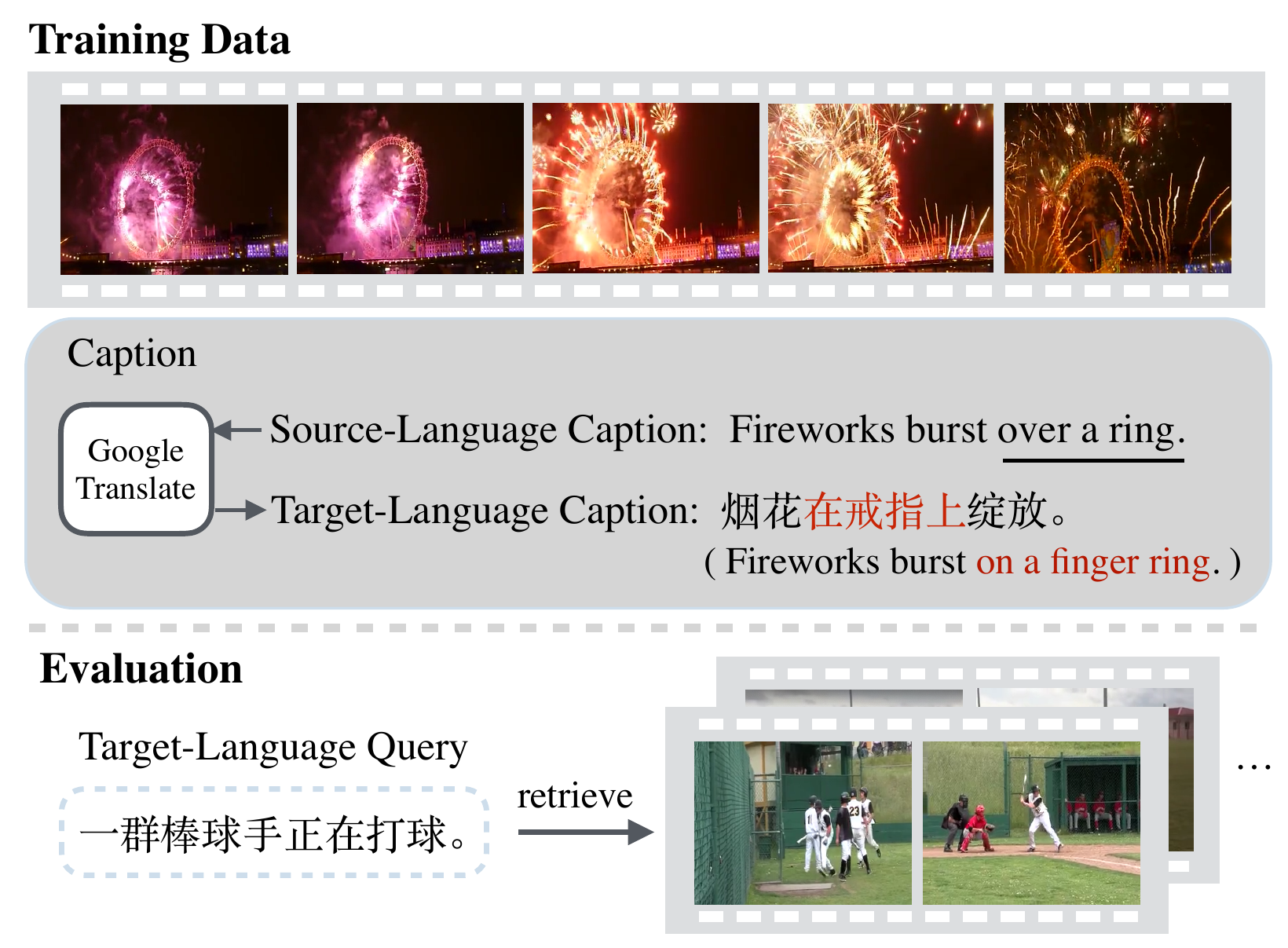}
\vspace{-4mm}
\caption{
A pipeline of cross-lingual cross-modal retrieval (CCR). A CCR model is trained on a collection of videos/images associated with captions in the source language (English). During evaluation, a query in the target language (Chinese) is given to retrieve relevant videos. 
}\label{fig:title-pic}
\vspace{-4mm}
\end{figure}

With the rapid  emergence of images and videos on the Internet such as on Facebook and TikTok, it has
brought great challenges to retrieve multimedia contents accurately~\cite{wei2019neural,liu2022temporal,xiao2021boundary,liu2021hierarchical}.
Recently, the task of cross-modal retrieval \cite{li2021value,liu2021progressive,Miech_2021_CVPR,cao2022visual,gabeur2022masking, Ali_2022_WACV,lu2019online,wang2022many,cui2019scalable,yang2020tree} has received growing research attention.
The majority of research work on cross-modal retrieval is dedicated to the English language, due to the availability of a large quantity of human-labeled data. 
With this regard, cross-lingual cross-modal retrieval (CCR), especially the one transferring from the source language with rich resources (e.g., English) to the target language where the human-labeled data is scarce or even not available, is of great importance.
Figure~\ref{fig:title-pic} shows a pipeline of CCR, where human-labeled target-language data is not provided during training.

Although many recent works~\cite{yu2021assorted, elliott2017imagination,wehrmann2019language,kim2020mule, burns2020learning, song2021product, lei2021mtvr,li2016adding} have achieved remarkable progress in multilingual scenarios, 
their approaches cannot be applied to CCR directly, since they require large-scale labeled datasets covering all interested languages.
Unfortunately, existing human-labeled resources for low-resource languages (e.g., Czech) are rather limited, and it is extremely expensive and time-consuming to manually annotate videos (or images) with descriptions in multiple languages.
However, with MT gaining popularity, research on MT is constantly growing recently \cite{park2021should, philip2021revisiting, cui2021directqe} and has shown the potential to overcome above problem.
In this paper, we focus on cross-lingual cross-modal retrieval utilizing MT.

To the best of our knowledge, only a few recent works \cite{portaz2019image, aggarwal2020towards, ni2021m3p, zhou2021uc2, huang2021multilingual} have paid attention to cross-modal retrieval in the cross-lingual setting.
Among them, Portaz \etal \cite{portaz2019image} and Aggarwal \etal \cite{aggarwal2020towards}
heavily rely on pre-trained multilingual word embeddings and pre-trained sentence encoders, respectively.
Instead, MMP~\cite{huang2021multilingual}, M3P ~\cite{ni2021m3p} and UC$^2$~\cite{zhou2021uc2} perform multilingual pre-training on their own, among which UC$^2$ constructs a multilingual vision and language (V+L) dataset with MT for 
pre-training and achieves the best performance.
Although they have proved that multilingual pre-training is beneficial to CCR, they only use source-language data for fine-tuning in the cross-lingual setting, which hurts the 
cross-lingual ability of models.
Besides, we empirically observe that outputs of MT are far from being perfect, and things get worse when input sentences are complicated, which has been ignored in previous works.
As exemplified in Figure~\ref{fig:title-pic}, the red Chinese words are the incorrect translation of underlined English words, where translation noise is introduced when translating the English caption into Chinese using Google Translate. 
Due to the existence of such noise, these machine-translated sentences are usually weakly-correlated with the corresponding visual content. 
Straightforward application of 
cross-modal matching to weakly-correlated data will result in corrupted representations and yield deteriorated performance, as neural models have a strong capacity to fit to the given (noisy) data.

To conquer the obstacle of exploiting machine-translated sentences, we propose noise-robust learning in this paper. 
Instead of revising the translated sentences to improve their quality, we introduce the noise-robust representation learning based on multi-view self-distillation to generate soft pseudo-targets. 
Specifically, we employ the cross-attention in Transformer to gather information from tokens that are more likely to be correctly translated according to the source-language counterparts, and filter out others. 
The output of the cross-attention is not only aligned with the source language, but also relatively clean than only using translated sentences. 
Thus we use the representation in the cross-attention module to generate the soft pseudo-targets, which provides direct supervision for the target-language encoding. 
Furthermore, inspired by the back-translation in unsupervised MT~\cite{artetxe2017unsupervised, huang2021comparison}, we utilize the cycle semantic consistency to minimize the semantic discrepancies between source sentences and back-translated sentences to further improve the noise robustness of the textual
encoder.

The main contributions of this paper are summarized in four aspects: 1) We propose noise-robust learning to solve the noise problem caused by MT for CCR. By resorting to MT, we conduct cross-lingual transfer without relying on the manually annotated target-language data. To the best of our knowledge, this work is the ﬁrst work focusing on the noise problem caused by MT for CCR.
2) To reduce the impact of translation noise, we design a novel noise-robust learning framework that introduces the multi-view distillation with cross attention to generate soft pseudo-targets and provide direct supervision to learn the robust vision-target language matching.
Further, we utilize the cycle semantic consistency to strengthen the capability of extracting semantic information from noisy sentences. 3) We construct a bilingual dataset MSR-VTT-CN, a new cross-lingual video-text retrieval dataset extending MSR-VTT with manually translated Chinese sentences for evaluation. 4) Extensive experiments are conducted on two cross-lingual video-text retrieval datasets, \ie VATEX and MSR-VTT-CN, and a cross-lingual image-text retrieval dataset Multi-30K, where our proposed model achieves a new state-of-the-art.

\section{Related Work} \label{sec:rel-work}

\subsection{Cross-modal Retrieval}

\textbf{Video-Text Retrieval.} The majority of recent works on video-text retrieval have focused on two types of approaches, namely concept-based approaches~\cite{dalton2013zero, chang2015semantic, markatopoulou2017query, habibian2016video2vec} and embedding-based approaches~\cite{song2019polysemous, gabeur2020multi, han2021fine, yang2021deconfounded,liu2021progressive, wu2021hanet, luo2021coco, wei2021meta,song2021spatial,wang2022siamese}. Among them, embedding-based approaches
currently dominate video-text retrieval, which measure similarities between text and videos by projecting them into single
or multiple common spaces. For instance, Gabeur \etal \cite{gabeur2020multi} design a multi-modal Transformer for video encoding and use BERT~\cite{devlin-etal-2019-bert} to obtain embeddings for text.
Besides, some recent models \cite{huang2017instance, zhu2020actbert, luo2021clip4clip, lei2021less} convert the text-to-video 
retrieval into a binary classification task, where models predict if text and videos match or not based on the fusion of their embeddings.

\textbf{Image-Text Retrieval.} Similar to video-text retrieval, most existing methods for image-text retrieval~\cite{liu2017learning, wang2018learning, Li_2019_ICCV, zeng2021conceptual, Anwaar_2021_WACV, he2021cross, gan2020large} aim to learn image and text features in a shared embedding space where similarities can be easily calculated. 
For example, Anwaar \etal \cite{Anwaar_2021_WACV} introduce an auto-encoder-based model to learn the 
composed representation of images and text queries, which are  mapped to the target image space to learn a similarity metric. 
Recently, there are some pre-trained models \cite{hong2021gilbert, li2020unicoder, li2020oscar} seeking to learn vision-language joint representations based on self-attention or cross-attention mechanism. 
These methods take a deep dive into the interaction between different modalities, 
and demonstrate advantageous retrieval performances.

Despite of performance improvements brought by above methods,
their models are designed and trained in a monolingual setting, 
and therefore lack the cross-lingual ability.

\begin{figure*}[tb!]
\centering\includegraphics[width=1.9\columnwidth]{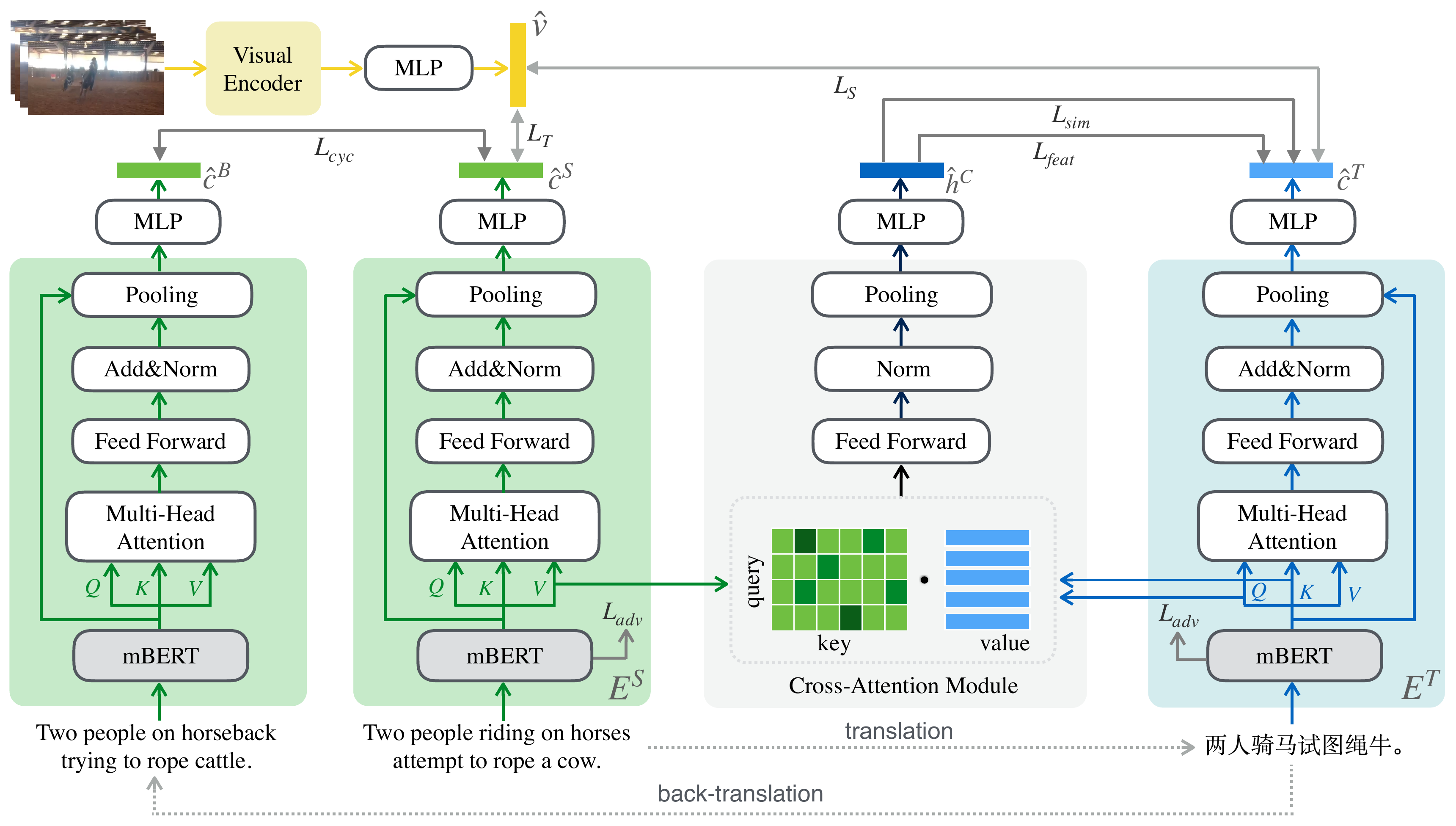}
\vspace{-4mm}
\caption{Illustration of the proposed NRCCR method for CCR. 
The source-language branch $E^S$ and target-language branch $E^T$ make up of the \textit{Basic Model} which aligns the video (or image) with source- and target-language input sentences, and a cross-attention module is adopted for noise-robust representation learning via self-distillation. The leftmost branch encodes the back-translated sentence which is supervised by $E^S$ with cycle semantic consistency. All textual branches is built upon pretrained multilingual BERT, which is trained to learn language-agnostic representations in an adversarial fashion. 
 }\label{fig:framework}
 \vspace{-3mm}
\end{figure*}

\subsection{Cross-lingual Cross-modal Retrieval}

Early works~\cite{portaz2019image, aggarwal2020towards}
on CCR adapt existing vision-language methods to the cross-lingual setting. 
Portaz \etal \cite{portaz2019image} resorts to non-contextualized multilingual word embeddings (BIVEC~\cite{luong2015bilingual} and MUSE~\cite{conneau2017word}), which are pre-trained by aligning words in different languages. 
Aggarwal \etal \cite{aggarwal2020towards} rely on pre-trained sentence encoders (mUSE ~\cite{yang2019multilingual} and LASER~\cite{artetxe2019massively}) to embed sentences from different languages into a common space.
Recently, with the release of large-scale multi-modal datasets~\cite{jia2021scaling, sharma2018conceptual}, some V+L pretraining models \cite{ni2021m3p, huang2021multilingual, zhou2021uc2,fei2021cross} are proposed to further narrow the 
gap between different languages and modalities.
Fei \etal \cite{fei2021cross} propose a cross-lingual cross-modal pretraining framework, which can be used for the downstream monolingual cross-modal retrieval task by finetuning on manually annotated target-language data. 
Ni \etal \cite{ni2021m3p} propose a Transformer-based encoder and two pre-training objectives to learn explicit alignments between vision and different languages, which use a multilingual-monomodal corpus and a monolingual-multimodal corpus to alleviate the issue of lacking enough non-English labeled data. 
Following \cite{ni2021m3p}, Huang \etal \cite{huang2021multilingual} design a Transformer-based pre-trained model MMP for learning contextualized multilingual multi-modal representations, which are pre-trained using Multi-HowTo100M.
Considering the fact that large-scale multi-modal datasets tend to heavily skewed towards well-resourced languages,
 the scarcity problem of aligned multilingual data remains a great challenge. To solve this problem, Zhou \etal \cite{zhou2021uc2} propose a MT-augmented cross-lingual cross-modal pre-training framework UC$^2$ and construct a multilingual V+L corpus.
 Although this work has made a breakthrough in CCR, it ignores the fact that translation results tend to be quite noisy, which would inevitably hurt model performances on  low-resource languages. 
 To alleviate this, in this paper, we focus on improving the robustness of CCR model against noise brought by MT.

\section{Methods}
\label{sec:method}

In this section, we present our \textbf{Noise-Robust Cross-lingual Cross-modal Retrieval (NRCCR)} model.
As shown in Figure~\ref{fig:framework}, it includes a source-language branch ($E^S$) and a target-language branch ($E^T$) responsible for 
source- and target-language text encoding, respectively.
Besides, our model is equipped with a cross-attention module, which plays a key role in filtering out translation noise.
To further improve its robustness, we use an additional branch (the leftmost branch in Figure~\ref{fig:framework}) to
encode back-translated sentences and then perform cycle semantic consistency learning. 
Adversarial training is also introduced to extract language-agnostic features.
In what follows, we first introduce the task definition and a basic model for CCR, followed by the description of our proposed model.

\subsection{Task Definition}
We first formally define the setting of cross-lingual cross-modal retrieval (CCR).
It involves two languages, namely
the source language $S$ and the target language $T$.
For the the source language $S$, we have a collection of human-labeled
training data $\mathcal{D}^S = \{d_1,d_2, ... , d_n\}$, where each instance $d_i$ consists of a caption $s^S_i$ paired with an image or video $v_i$.
As for the target language $T$, due to the scarcity of human-labeled
data,  we assume there are no resources for it but some external tools,
such as Google Translate.
The core task of CCR is to obtain a model applicable in the target language $T$, without using any manually annotated target-language data.

\subsection{A Basic Model for CCR}
\label{sec:two-branch}
\noindent \textbf{Input data.}
Since human-labeled training data is only available for the source language, we automatically translate the gold-standard source dataset $\mathcal{D}^S$ into the target language.
Concretely, for each instance in $\mathcal{D}^S$, we translate the caption $s^S$ into a target-language sentence $s^T$, utilizing MT\footnote{In this paper, we use Google Translate (\url{https://translate.google.cn}).}.
In this way, we construct a collection of triplet $\mathcal{D}^T = \{I_1,I_2, ... , I_{n}\}$, where each instance $I_i = (v_i, s^S_i, s^T_i)$ consists of a video (or an image) $v_i$, a source-language caption $s^S_i$ and a translated target-language caption $s^T_i$.

\textbf{Visual encoder.}
Given a video $v$, we first extract a sequence of representations $U=\{u_1,u_2,\dots, u_l\}$ using pre-trained 2D CNN, where $u_i \in \mathbb{R}^{d_u}$ denotes the representation of $i$-th frame. 
Then, we feed it into a Transformer block, to enhance the  frame-inter interaction and generate a video feature vector $v \in \mathbb{R}^{d_v}$ :
\begin{equation}
    \begin{array}{cc}
    
        v = f(Transformer_v(U))
    
    \end{array}
\end{equation}
where $f(\cdot)$ denotes the operation of average pooling.

When input is an image, similar with \cite{portaz2019image, aggarwal2020towards}, we use the ResNet-152~\cite{xie2017aggregated} pretrained on ImageNet\cite{5206848} dataset as our backbone, and keep it frozen. 
We use the output of last average pooling layer in ResNet-152 into a trainable FC layer to obtain the image feature vector $e \in \mathbb{R}^{d_e}$.
In the following, for the ease of description, we take video-text retrieval as an example to depict our methods in detail.

\textbf{Textual encoder.}
We design a two-branch encoder for CCR, which consists of a source-language branch and a target-language branch. 
As illustrated in Figure~\ref{fig:framework}, each branch includes a Transformer block built upon the pre-trained multilingual BERT (mBERT \cite{devlin-etal-2019-bert}). 
To extract the shared information between different languages, both branches are weight-sharing. The inputs of them are pseudo-parallel sentence pairs $\{s^S, s^T\}$.
Specifically, given a source-language sentence $s^S$ and a target-language sentence $s^T$, we use mBERT to generate a sequence of multilingual representations $m^S \in \mathbb{R}^{N \times d_w}$ and $m^T \in \mathbb{R}^{M \times d_w}$ containing $N$ and $M$ tokens, respectively. 
Then, we feed $m^S$ and  $m^T$ to a high-level Transformer block to further extract task-specific features:

\begin{equation}
    \begin{array}{cc}
        c^S = f(Transformer_t(m^S)),
        c^T = f(Transformer_t(m^T))
    \end{array}
\end{equation}
where $c^S$ and $c^T$ denote the sentence embedding of $s^S$ and $s^T$. 
For the ease of reference, we term the source-language branch as $E^S$ and the target-language branch as $E^T$:
\begin{equation}
    \begin{array}{cc}
         c^S = E^S(s^S),
         & c^T = E^T(s^T)
    \end{array}
\end{equation}

\textbf{Mapping into a common space.}
To promote alignments of the sentence in different languages with the corresponding video $v$, we try to learn two mappings $g_t(\cdot)$ and $g_v(\cdot)$ to respectively project the different language representations and the video representation into a common space as follows:

\vspace{-1mm}
\begin{equation}
    \begin{array}{cc}
        \hat{c}^S = g_{t}({c^S}), 
        \hat{c}^T = g_{t}({c^T}), 
        \hat{v} = g_{v}(u)
    \end{array}
\end{equation}
where $\hat{c}^S $, $\hat{c}^T$ and $\hat{v}^T$ are source-language sentence embedding, target-language sentence embedding and video embedding, respectively. Besides, each mapping is implemented as a multi-layer perceptron.
To make relevant  video-sentence pairs near and irrelevant pairs far away in the common space, an improved triplet ranking loss~\cite{faghri2017vse++, cvpr2019-dual-dong} is employed which penalizes the model according to the hardest negative examples in the mini-batch:

\begin{equation}
\label{eq:triplet_loss}
    \begin{aligned}
        \mathcal{L}^S(\hat{c}^{S}, \hat{v}) & =  max(0, \triangle + sim(\hat{v}, \hat{c}^{S-}) - sim(\hat{v}, \hat{c}^{S})) \\ 
        & +  max(0, \triangle + sim(\hat{c}^{S}, \hat{v}^{-}) - sim(\hat{c}^{S}, \hat{v}))  \\
        \mathcal{L}^T(\hat{c}^{T}, \hat{v})  & =  max(0, \triangle + sim(\hat{v}, \hat{c}^{T-}) - sim(\hat{v}, \hat{c}^{T})) \\ 
         & + max(0, \triangle + sim(\hat{c}^{T}, \hat{v}^{-}) - sim(\hat{c}^{T}, \hat{v}))  \\
    \end{aligned}
\end{equation}
where $\triangle$ indicates a margin constant and $sim(\cdot)$ denotes the similarity function, \eg cosine similarity, while $\hat{c}^{S-}$, $\hat{c}^{T-}$and $\hat{v}^{-}$ respectively indicate the embedding of a hard negative source-language sentence sample, a hard target-language  sentence sample and a negative video sample. 
The overall triplet loss is computed as $\mathcal{L}_{tri} = \mathcal{L}^S + \alpha \mathcal{L}^T$, where $\alpha$ is a scalar whose 
value is set empirically.

\begin{figure}[tb!]
\centering\includegraphics[width=0.85\columnwidth]{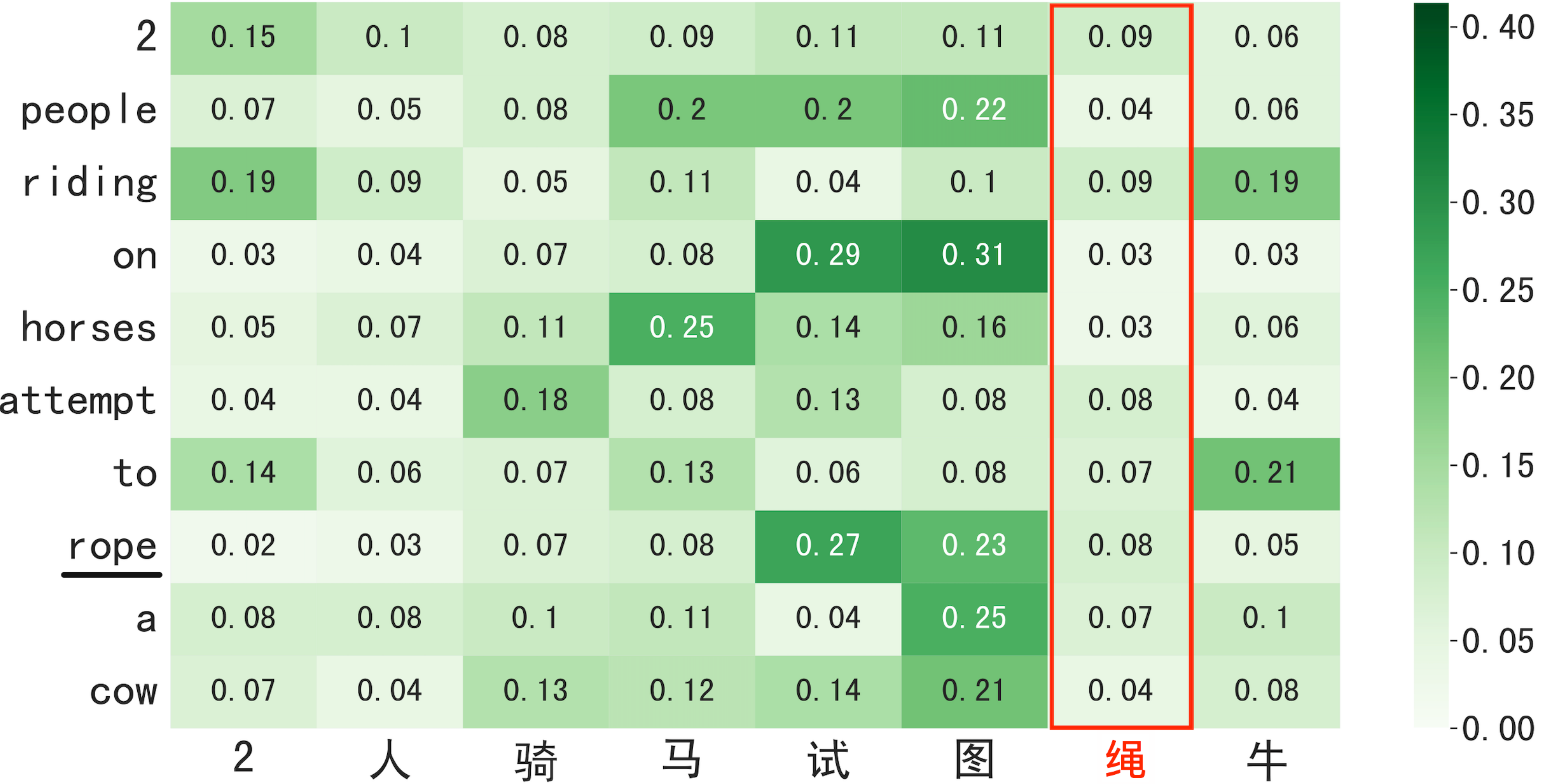}
\vspace{-2mm}
\caption{An example of the cross-attention map.
The red represents the incorrect translated word of the underlined source-language word. }\label{fig:cross-attention}
\vspace{-5mm}
\end{figure}

\subsection{Our Proposed NRCCR}

On the base of \textit{Basic Model}, we additionally introduce noise-robust representation learning, cycle semantic consistency and language-agnostic representation learning, resulting in our proposed NRCCR.

\subsubsection{\textbf{Noise-Robust Representation Learning}}
\label{sec:NRL}
In order to improve the noise robustness of our model, we devise a cross-attention module and multi-view self-distillation.

\textbf{Cross-attention module.}
Since pseudo-parallel sentence pairs are constructed with MT, this process would inevitably introduce translation noise, such as inaccurate translation and grammatical mistakes. 
However, in \textit{Basic Model}, no matter how noisy $s^T$ is, minimizing $\mathcal{L}^T$ always closes the distance between embeddings of $s^T$ and $v$, which would result in corrupted representations and thereby compromise the retrieval performance.
To address this, we propose to generate relatively clean target-language representations to guide the training of the target-language branch via  multi-view self-distillation. 
Therefore, we introduce a cross-attention module $CrossATT$ to collect information from tokens in $s^T$ that are more likely to be correctly translated according to the source-language counterparts, and filter out others.
The cross-attention is derived from self-attention, and the difference is that the input of cross-attention module comes from different languages.
Specifically, given a parallel sentence pair $(s^S, s^T)$, we first feed $s^S$ and $s^T$ into pre-trained mBERT and obtain their token-level representations $m^S$ and $m^T$ respectively.
The cross-attention module computes as follows:

\begin{equation}
\label{eq:cross}
    \begin{aligned}
        h &= CrossATT(m^S, m^T, m^T) \vspace{1ex}\\
        & = softmax(\frac{W_Q m^S \cdot {(W_K m^T})^T}{ \sqrt{d_w} })\cdot W_V m^T \vspace{1ex}\\
         h^C &= Norm(FFN(h))
    \end{aligned}
\end{equation}
where $W_Q, W_K$ and $W_V$ are three projection matrices for transforming the input features,
$FFN$ and $Norm$ indicate the feed-forward networks and  layer normalization operation in $Transformer_t$, respectively.
The output $h$ is calculated by multiplying value with attention weights, which comes from the similarity between the corresponding query and all keys.
As a result, among all tokens in $s^T$, the token that is more similar to the query of $s^S$ would hold a larger attention weight and contribute more to the output, otherwise, the cross-attention would assign a low weight to the dissimilar tokens which are noisy. 
Figure~\ref{fig:cross-attention} gives an example, where the verb "rope" is incorrectly translated as a noun "string" by Google Translate. Hence, the attention weights between the incorrect word (marked in red) and source-language words are assigned low values.
In this way, $CrossATT$ automatically gathers relatively clean information from the target-language sentence $s^T$, improving the quality of the output $h^C$.

\textbf{Multi-view self-distillation module.} 
To alleviate even eliminate the influence of the noise generated by MT, we design a multi-view self-distillation loss, which guides the training of target-language branch from the similarity-based view and feature-based view, respectively.

\textit{Similarity-based view: }
The key to cross-modal retrieval is measuring the similarity between different modalities. 
Here we regard the cross-attention module as a teacher. Its outputs take part in the calculation of cross-modal similarities, which are then used as pseudo-targets to supervise the target-language branch learning.
Specifically, we first project the representation $h^C$ into the common space, 
obtaining a sentence embedding expressed as $\hat{h}^C = g_{t}(f(h^C))$. \
Then, we calculate the softmax-normalized text-to-video similarity $q^{t2v}(\hat{h}^C ,\hat{v})$  and video-to-text similarity 
$q^{v2t}(\hat{v}, \hat{h}^C)$ as:

\begin{equation}
    \begin{array}{cc}
          q^{t2v}(\hat{h}^C, \hat{v}) 
          = \frac{exp(sim(\hat{h}^C,  \hat{v}) / \tau)}
          {\sum_{j=1}^{B}exp(sim(\hat{h}^C,  \hat{v}_j) / \tau)} \vspace{1.5ex}\\
          q^{v2t}(\hat{v}, \hat{h}^C) 
          = \frac{exp(sim(\hat{v}, \hat{h}^C) / \tau)}
          {\sum_{j=1}^{B}exp(sim(\hat{v}, \hat{h}^C_j) / \tau)} \\
    \end{array}
\end{equation}
where $B$ denotes the size of a minibatch from $\mathcal{D}^T$, and $\tau$ is the temperature coefficient.
For the target-language sentence $s^T$, we similarly compute the softmax-normalized text-to-video similarity $p^{t2v}(\hat{c}^T, \hat{v}) $ and video-to-text similarity $p^{v2t}(\hat{v}, \hat{c}^T)$ similarity.
Finally, the Kullback Leibler (KL) divergence is employed as the similarity-based distillation loss:

\begin{equation}
    \begin{aligned}
         \mathcal{L}_{sim} = \frac{1}{2} 
         [ &KL(q^{t2v}(\hat{h}^C ,\hat{v}) || p^{t2v}(\hat{c}^T, \hat{v}))\\ 
         + &KL(q^{v2t}(\hat{v}, \hat{h}^C)  || p^{v2t}(\hat{v}, \hat{c}^T))]
    \end{aligned}
\end{equation}

\textit{Feature-based view: }
Another supervision from the cross-attention module is on a feature-based view, where the output features of the teacher are  utilized to guide the student's learning process.
Here we use the feature-based loss $\mathcal{L}_{feat}$ to introduce inexplicit knowledge in the sentence embedding $\hat{h}^C$ to the encoding of  target-language sentence $s^T$.
As shown in Equation ~\ref{eq:l1},  $\mathcal{L}_{feat}$ is defined as  the $L1$ distance between  $\hat{h}^C$ and $\hat{c}^T$:
\begin{equation}
\label{eq:l1}
    \begin{array}{cc}\
         \mathcal{L}_{feat} =||f(\hat{h}^C) - f(\hat{c}^T)||_1
    \end{array}
\end{equation}

\subsubsection{\textbf{Cycle Semantic Consistency}}
\label{sec:csc}
As we would use source-language sentences generated by MT as the source-branch input during inference, it is necessary to enhance the noise robustness of our model on the source language.
To this end, inspired by the back-translation applied to unsupervised
machine translation~\cite{artetxe2017unsupervised, huang2021comparison}, we propose  to improve the cycle semantic consistency between sentences and their back-translated counterparts.
Concretely, given a source-language sentence $s^S$, we first utilize  MT  to translate it into the target language and then back again to obtain a source-language sentence $s^B$, \ie $s^B = BackTranslation(s^S)$.
For sentence $s^S$, the source-language branch $E^S$ is expected to extract consistent semantic information from the back-translated sentence $s^B$ with original sentence $s^S$ in the translation cycle, \ie $E^S(s^S) \approx E^S(s^B)$.
Similar with Equation ~\ref{eq:triplet_loss}, we can incentivize this behavior using an improved triplet ranking loss:
\begin{equation}
    \begin{aligned}
         \mathcal{L}_{cyc}(s^S,s^B) &= max(0, m + sim(\hat{c}^{B}, \hat{c}^{S-}) - sim(\hat{c}^{B}, \hat{c}^{S})) \\
         & + max(0, m + sim(\hat{c}^{S}, \hat{c}^{B-}) - sim(\hat{c}^{S}, \hat{c}^{B}))
    \end{aligned}
\end{equation}
where $\hat{c}^{B} =  E^S(s^B)$.
Although $s^S$ and $s^B$ are in the same language, $s^B$ is even noisier 
than $s^T$ as translation noise tends to be amplified during back-translation. 
Therefore, minimizing $\mathcal{L}_{cyc}$ can further 
improve the robustness of our model to against noise
and produce semantic consistency features.
Besides, back-translation can be also thought of as a data-augmentation technique, which could improve the diversity of words.

\subsubsection{\textbf{Language-agnostic Representation Learning}}
\label{sec:adv}
Considering language-specific features lack the cross-lingual transfer ability,
we train the textual encoder in an adversarial fashion to generate language-agnostic representations. 
Specifically, we construct a language classifier $F$ to act as the discriminator and is adopted to distinguish the source and target languages. 
The language classifier is consisted of a multi-layer feed-forward neural networks, and the discriminator loss in adversarial training is defined as:
\begin{equation}
    \mathcal{L}_{adv} =  \frac{1}{|\mathcal{D}^T|}
    \sum \limits_{I_i \in \mathcal{D}^T}
    [log F(f(m^S_i)) +
    log(1 - F(f(m^T_i)))]
\end{equation}
The text encoder $E^S$ and $E^T$ aim to generate language-agnostic representations and confuse the discriminator $F$, while discriminator $F$ aims to distinguish between target-language sentences and source-language sentences.

\subsubsection{\textbf{Training and Inference}}
\label{sec:TrainInference}
Our model is trained by minimizing the combination of the above losses.
To sum up, the total loss function is defined as:
\begin{equation}
    \begin{array}{cc}
       \mathcal{L} =  \mathcal{L}_{tri} + \lambda_1 \mathcal{L}_{sim} + \lambda_2 \mathcal{L}_{feat} +\lambda_3 \mathcal{L}_{cyc} - \lambda_4 \mathcal{L}_{adv}
      \label{eq:all}
    \end{array}
\end{equation}
where $\lambda_1, \lambda_2, \lambda_3, \lambda_4$ are all hyper-parameters to balance the importance of each loss.

After the model has been trained, given a sentence query in the target language, we sort candidate videos/images in descending order according to their cross-modal similarity with the query.
To compute the similarity between a video (or an image) $v$ and a sentence query in the target language $s^T$, we first translate $s^T$ to the source-language sentence $s^S$. The similarity between $v$ and $s^T$ is computed as the sum of their similarity and the corresponding similarity between $v$ and the translated sentence $s^S$ from $s^T$. Formally, the final similarity is computed as:
\begin{equation}
    \begin{aligned}
        Score(v,s^T) = \beta  sim(\hat{v}, \hat{c}^T)
        + (1 - \beta) sim(\hat{v}, \hat{c}^S) \label{eq:beta}
    \end{aligned}
\end{equation}
where $\beta$ is a empirically-set scalar weight.

\section{Experiment} \label{sec:eval}

\begin{table*} [tb!]
\renewcommand{\arraystretch}{1.2}
\vspace{-3mm}
\caption{Performance comparison of cross-lingual video-text retrieval on VATEX (the source language is English and the target language is Chinese).
Symbol asterisk (*) indicates the model is pre-trained on Multi-HowTo100M \cite{huang2021multilingual}.
}
\label{tab:sota-vatex}
\centering 
\scalebox{0.99}{
\begin{tabular}{@{}l*{12}{r}c @{}}
\toprule
\multirow{2}{*}{\textbf{Method}}   & \multicolumn{5}{c}{\textbf{Text-to-Video Retrieval}} && \multicolumn{5}{c}{\textbf{Video-to-Text Retrieval}} & \multirow{2}{*}{\textbf{SumR}} \\
 \cmidrule{2-6}  \cmidrule{8-12} 
& R@1 & R@5 & R@10 & Med r & mAP && R@1 & R@5 & R@10 & Med r & mAP & \\
\cmidrule{1-13}
MMP w/o pre-train\cite{huang2021multilingual}  &23.9 &55.1 &67.8 &- &- && - &- &- &- &- &-\\
MMP \cite{huang2021multilingual}* &29.7 &63.2 &\textbf{75.5} &- &- && - &- &- &- &- &- \\
\cmidrule{1-13}
W2VV \cite{8353472}   &5.46  &12.3  &15.4  & 298 &9.20  && - & - & - & - & - & - \\
VSE++ \cite{faghri2017vse++}  &21.1 & 48.1 &59.6 & 6.0 &33.72 && 34.9 &67.2 &77.5 &3.0 &21.76 & 307.7\\
W2VV++ \cite{li2019w2vv++} &20.5 &48.3 &59.5 &6.0 &33.40 && - & - & - & - & - & - \\
Miech \etal \cite{miech19howto100m} &13.7 &37.1 &50.4 &10.0 &25.32 && 23.2 &53.1 &67.3 &5.0 &14.62 &244.8\\
CE \cite{liu2019use} &20.2 &48.5 &60.8 &6.0 &33.48 && 31 &63.5 &76.7 &3.0 &20.2 & 300.7\\
Dual Encoding \cite{cvpr2019-dual-dong} &23.1 &52.1 &62.6 &5.0 &36.32 && 35.6 & 67.9 & 79.3 & 3.0 &23.98 & 320.5 \\
HGR \cite{han2021fine}  &14.3 &37.0 &47.5 &12.0 &25.10 && 27.8 &61.0 &72.9 &4.0 &15.5 & 261.0\\
GPO \cite{chen2021learning} &19.5 &46.9 &57.8 &6.0 &32.20 && 33.8 &65.9 &77.3 &3.0 &20.21 & 301.3\\
RIVRL \cite{dong2022reading} &26.8 &56.5 &67.3 &4.0 &40.35 && 38.3 &71.1 &\textbf{80.9} &\textbf{2.0} &26.62 & 340.9\\ 
\cmidrule{1-13}
{NRCCR}  & \textbf{30.4}  & \textbf{65.0}  &{75.1}  &\textbf{3.0}  &\textbf{45.64} && \textbf{40.6}  &\textbf{72.7}  &\textbf{80.9}  &\textbf{2.0}  &\textbf{32.40} & \textbf{364.7} \\
[3pt]
\bottomrule
\end{tabular}
 }
\end{table*}

\subsection{Experimental Settings} \label{ssec:exp-set}

\textbf{Datasets.}
We conduct experiments on two public multilingual cross-modal retrieval datasets (VATEX~\cite{wang2019vatex} and Multi30K~\cite{elliott2016multi30k}) and MSR-VTT-CN constructed by ourselves.

\begin{itemize}[leftmargin=*]
\item \textbf{VATEX}~\cite{wang2019vatex}: VATEX is a bilingual video-caption dataset, which contains over 41,250 videos and 825,000 captions. 
Each video  corresponds to 10 English sentences and 10 Chinese sentences describing the video content.
Following \cite{dong2021dual, chen2020fine}, we use 25,991 video clips for training, 1,500 clips for validation and 1,500 clips for test.
\item \textbf{Multi30K}~\cite{elliott2016multi30k}: 
This dataset contains 31K images, which is built upon Flickr30K \cite{young2014image}. It provides five captions per image in English and German and one caption per image in French and Czech. Following Flickr30K \cite{young2014image}, we split the data 29000/1014/1000 as train/dev/test sets.
\item \textbf{MSR-VTT-CN}: MSR-VTT \cite{xu2016msr} is a monolingual dataset which has been widely used for video-text retrieval. 
We build MSR-VTT-CN  by extending  MSR-VTT to a bilingual version, which follows the partition of \cite{yu2018joint} containing 9,000 and 1,000 for the training and testing, respectively. 
For each video in the training set of MSR-VTT, we employ Google Translate to automatically translate its captions from English to Chinese. 
MSR-VTT-CN test set is obtained by manually translating the test set of MSR-VTT into Chinese. 
To ensure the quality of manual-annotated data, similar to \cite{lan2017fluency}, we hire ten Chinese native speakers who are also good at English (passed the National College English Test 6).
Since some English captions are ambiguous, 
native speakers also take video content into consideration to aid translation.
\end{itemize}

\begin{table} [tb!]
\renewcommand{\arraystretch}{1.2}
\setlength{\abovecaptionskip}{1.1mm} 
\setlength{\belowcaptionskip}{-0.1cm}
\caption{Performance comparison on MSR-VTT-CN. }
\label{tab:sota-msrvtt}
\centering 
\scalebox{0.95}{
\begin{tabular}{@{}l*{12}{r}c @{}}
\toprule
\multirow{2}{*}{\textbf{Method}}   & \multicolumn{2}{c}{\textbf{T2V}} && \multicolumn{2}{c}{\textbf{V2T}} & \multirow{2}{*}{\textbf{SumR}} \\
\cmidrule{2-3}  \cmidrule{5-6}
& R@10 & mAP && R@10 & mAP & \\
\cmidrule{1-7}
W2VV \cite{8353472} &22.9 &1.50 && - & - & -\\
VSE++ \cite{faghri2017vse++}&54.0 &29.57 &&55.0 &29.86 & 230.8\\
Miech \etal \cite{miech19howto100m} &52.9 &25.95 &&48.9 &25.31 &203.0\\
Dual Encoding \cite{cvpr2019-dual-dong} &56.6 &31.75 &&57.1 &32.25 &244.1\\
CE \cite{liu2019use} &63.4 &- &&62.7 &- & 265.4 \\
SEA \cite{li_2020sea} &61.1 &33.80 &&42.4 &21.90 &215.8 \\
HGR \cite{han2021fine} &53.3 &27.18 &&53.3 &27.79 & 219.5 \\
GPO \cite{chen2021learning} &53.2 &29.86 &&52.9 &29.15 & 226.5\\
RIVRL \cite{dong2022reading} & 63.0 & 37.03 && 63.5 & 35.46 &273.7 \\
\cmidrule{1-7}
NRCCR & \textbf{67.4} & \textbf{41.93} && \textbf{67.2} & \textbf{43.00} & \textbf{307.3} \\
[3pt]
\bottomrule
\end{tabular}
}
\end{table}

\textbf{Evaluation metrics.}
Following the previous works~\cite{dong2022reading,li2019w2vv++}, we use rank-based metrics, namely $R@K$ ($K = 1, 5, 10$), Median rank (Med r) and mean Average Precision (mAP) to evaluate the performance. $R@K$ is the fraction of queries that correctly retrieve desired items in the top $K$ of ranking list. Med r is the median rank of the first relevant item in the search results. Higher $R@K$, mAP and lower Med r mean better performance. 
For overall comparison, we report the Sum of all Recalls (SumR).

\begin{table*} [tb!]
\renewcommand{\arraystretch}{1.2}
\setlength{\abovecaptionskip}{1.5mm} 
\setlength{\belowcaptionskip}{-0.3cm}
\caption{Performance comparison of cross-lingual image-text retrieval on Multi-30K. 
}
\label{tab:sota-multi30k}
\centering 
\scalebox{1.0}{
\begin{tabular}{@{}l*{12}{r}c @{}}
\toprule
\multirow{3}{*}{\textbf{Method}}   & \multicolumn{3}{c}{\textbf{English -> German}} && \multicolumn{3}{c}{\textbf{English -> French}} &&
\multicolumn{3}{c}{\textbf{English -> Czech}}\\
 \cmidrule{2-4}  \cmidrule{6-8} \cmidrule{10-12}
 &\makecell[c]{T2I} &\makecell[c]{I2T} &\multirow{2}{*}{\textbf{SumR}}
 &&\makecell[c]{T2I} &\makecell[c]{I2T} &\multirow{2}{*}{\textbf{SumR}}
&&\makecell[c]{T2I} &\makecell[c]{I2T} &\multirow{2}{*}{\textbf{SumR}}
 \\
 \cmidrule{2-2}  \cmidrule{3-3}  \cmidrule{6-6} \cmidrule{7-7} \cmidrule{10-10} \cmidrule{11-11}
&R@10 & R@10 &&& R@10 & R@10 &&& R@10 & R@10\\
\hline
\textbf{Backbone with frozen ResNet152:}\\
ISUM \cite{portaz2019image} &\makecell[c]{44.20} & \makecell[c]{-} &{-} && \makecell[c]{46.00} &\makecell[c]{-} &{-}   &&\makecell[c]{-} & \makecell[c]{-} &{-}\\
TZClIR \cite{aggarwal2020towards} &\makecell[c]{48.80} &\makecell[c]{-} &{-} &&\makecell[c]{54.80} &\makecell[c]{-} &{-} &&\makecell[c]{-} & \makecell[c]{-} &{-}\\
{NRCCR} &\black{\makecell[c]{68.80}} &\black{\makecell[c]{67.30}}  &\black{306.8} &&\black{\makecell[c]{67.40}} &\black{\makecell[c]{69.30}}  &306.1 && \black{\makecell[c]{67.00}} &\black{\makecell[c]{66.20}}  &\black{301.3}\\ 
\hline
\textbf{Pre-trained on large-scale V+L datasets:}\\
MMP \cite{huang2021multilingual} &\makecell[c]{72.60} &\makecell[c]{-} &- &&\makecell[c]{-} &\makecell[c]{-} &\makecell[c]{-} &&\makecell[c]{78.20} &\makecell[c]{-} &-\\
M$^3$P \cite{ni2021m3p} &\makecell[c]{-} &\makecell[c]{-} &351.0 &&\makecell[c]{-} &\makecell[c]{-} &276.0 &&\makecell[c]{-} &\makecell[c]{-} &220.8\\
UC$^2$ \cite{zhou2021uc2} &\makecell[c]{-} &\makecell[c]{-} &449.4 && \makecell[c]{-} &\makecell[c]{-} &444.0 &&\makecell[c]{-} &\makecell[c]{-} &407.4\\
CLIP~\cite{radford2021learning} + Ours &\makecell[c]{\textbf{88.90}} &\makecell[c]{\textbf{89.20}} &\textbf{450.8} && \makecell[c]{\textbf{89.2}} &\makecell[c]{\textbf{89.7}} &\textbf{452.9} &&\makecell[c]{\textbf{87.9}} &\makecell[c]{\textbf{87.8}} &\textbf{\textbf{438.7}}\\
[3pt]
\bottomrule
\end{tabular}
 }
\end{table*}

\subsection{Performance Comparison} \label{ssec:exp-sota}

\subsubsection{\textbf{Cross-lingual Video-Text Retrieval}}
To the best of our knowledge, MMP~\cite{huang2021multilingual} is the only work specifically designed for cross-lingual video-text retrieval.
To further prove the effectiveness of our mode, following \cite{zhou2021uc2},we make the comparison with the traditional monolingual video-text retrieval methods by translating target-language testing samples into the source language with MT, which are typically regarded as the baselines for cross-lingual cross-modal retrieval.
Specifically, we compare following monolingual video-text retrieval methods including W2VV ~\cite{8353472}, VSE++~\cite{faghri2017vse++}, W2VV++~\cite{li2019w2vv++}, Miech \etal \cite{miech19howto100m}, CE \cite{liu2019use}, Dual Encoding~\cite{cvpr2019-dual-dong}, HGR \cite{han2021fine}, GPO \cite{chen2021learning}, and RIVRL ~\cite{dong2022reading}, considering their source code available.

\textbf{Experiments on VATEX.} 
In the cross-lingual setting, we use  VATEX English training set as the source-language labeled data,
and VATEX Chinese test set for target-language evaluation.
As shown in Table \ref{tab:sota-vatex}, our method outperforms MMP without pre-training with a clear margin, and achieves comparable results with MMP which is pre-trained on a large-scale dataset Multi-HowTo100M  \cite{huang2021multilingual} (the multilingual version of Howto100M~\cite{miech19howto100m}).
As for the adapted monolingual models, their performances are all overwhelmed by our model. 
In addition, we also try replacing Google MT with Baidu MT, and achieve a better result (SumR=375.2). 
These result demonstrates the compatibility of our model with different MTs.

\textbf{Experiments on MSR-VTT-CN.}
Similarly, in Table~\ref{tab:sota-msrvtt}, we report experimental results on MSR-VTT-CN.
Although models for comparison are all carefully designed and could achieve high performances when test input is clean, they still are no match for our model.
Since the SOTA method MMP \cite{huang2021multilingual} only performs experiments on VATEX and remains closed-source, its performance on MSR-VTT-CN is not reported.

\subsubsection{\textbf{Cross-lingual Image-Text Retrieval}}
To verify the cross-lingual transfer ability of our model for text-image retrieval, we conduct experiments on Mutlti30K. 
As can be seen in Table~\ref{tab:sota-multi30k}, SOTA methods are divided into two groups.
The first group includes methods that use the same image backbone (frozen ResNet152). 
ISUM \cite{portaz2019image} and TZCIR \cite{aggarwal2020towards} 
rely on multilingual word embeddings
and sentence encoders to obtain multilingual sentence representations, which are then simply mapped into a common space  with corresponding image representations.
Results in Table~\ref{tab:sota-multi30k} show that NRCCR consistently outperforms them with a clear margin.

The second group includes models performing large-scale V+L pertaining. 
Among them, MMP \cite{huang2021multilingual} is pre-trained on multilingual Howto100M, M$^3$P \cite{ni2021m3p} is pre-trained on Wikipedia and Conceptual Captions \cite{sharma2018conceptual}, and UC$^2$ \cite{zhou2021uc2} is pre-trained on multilingual Conceptual Captions which was extended by MT. 
\black{
We use an open-source pre-trained model CLIP~\cite{radford2021learning} which is pre-trained on Wikipedia
as the image encoder.
}
With the help of CLIP, our model achieves a significant improvement on all languages in this dataset. 
On French and Czech, our model significantly outperforms the latest-proposed method UC$^2$, which also resorts to MT but ignores the fact that translation results are noisy.
These results not only verify the effectiveness of noise-robust learning on multiple languages, but also demonstrate its compatibility with
stronger visual encoders.

\subsection{ Robustness Analysis}
Due to the usage of MT during training and test, translation noise is introduced in both stages.
In this section, we analyze the robustness of our model against translation noise during training and test, respectively. 
For a fair comparison, we compare our model to \textit{Basic Model}, which can be regarded as our NRCCR without noise-robust learning modules.

\begin{figure}[tb!]
\centering\includegraphics[width=0.99\columnwidth]{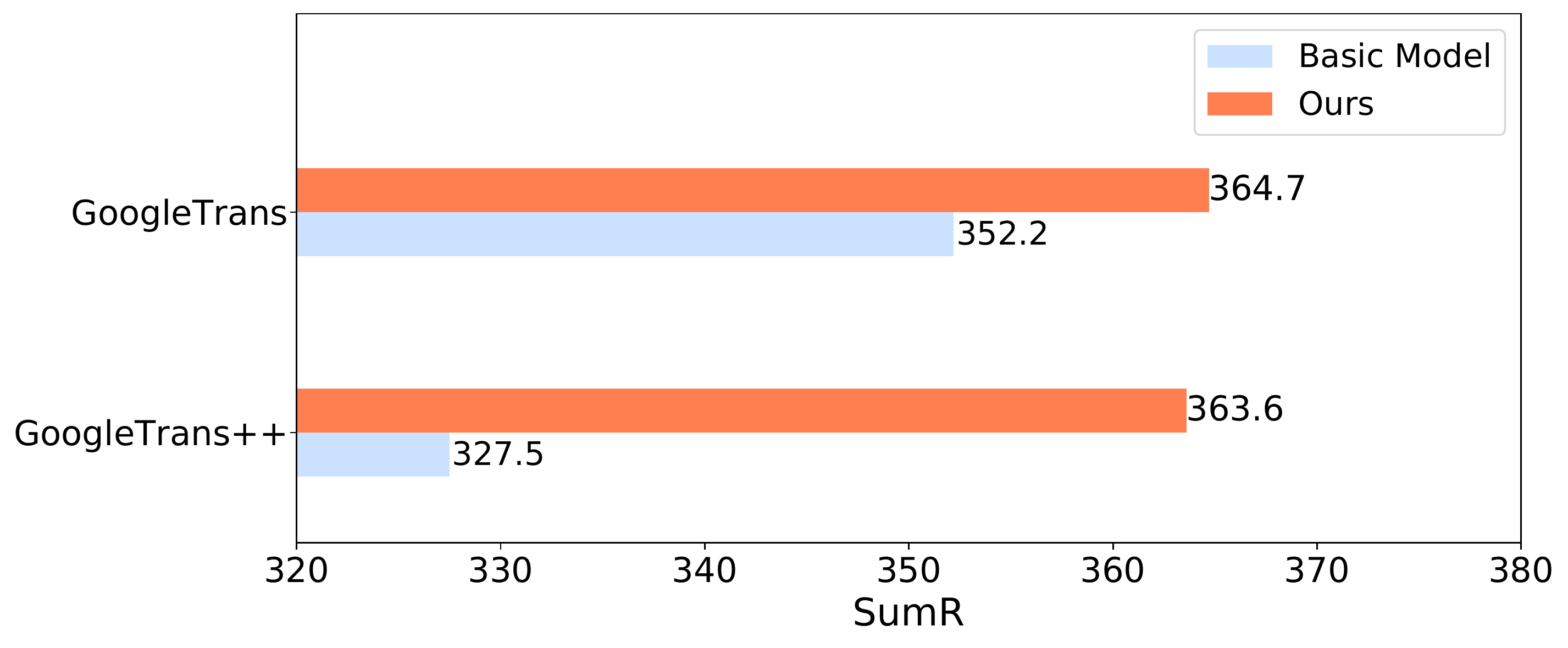}
\vspace{-3mm}
\caption{
Performances comparison on various degrees of noise during training. 
\textit{GoogleTrans++} means using Google Translate two more times, showing more noisy training data.
}\label{fig:noise-MT}
\end{figure}

\textbf{Noise robustness during training.}
To further prove the robustness of our model against translation noise, we conduct the experiment with various degrees of training noise on VATEX.
We evaluate models in the following two scenarios:
Training models with data obtained by performing English-to-Chinese MT  (EN$\rightarrow$ZH);
Training models with data obtained by applying MT two more times (EN$\rightarrow$ZH$\rightarrow$EN$\rightarrow$ZH).
For ease of reference, we term the former as \textit{GoogleTrans}, and the latter as \textit{GoogleTrans++}.
We argue that the training data used in the latter are noisier. 
As shown in Figure~\ref{fig:noise-MT}, after multiple-time translation being used, the performance of \textit{Basic Model} drops severely.
By contrast, the performance of our model is more stable, which verifies the effectiveness of our noise-robust representation learning.

\begin{figure}[tb!]
\centering\includegraphics[width=0.99\columnwidth]{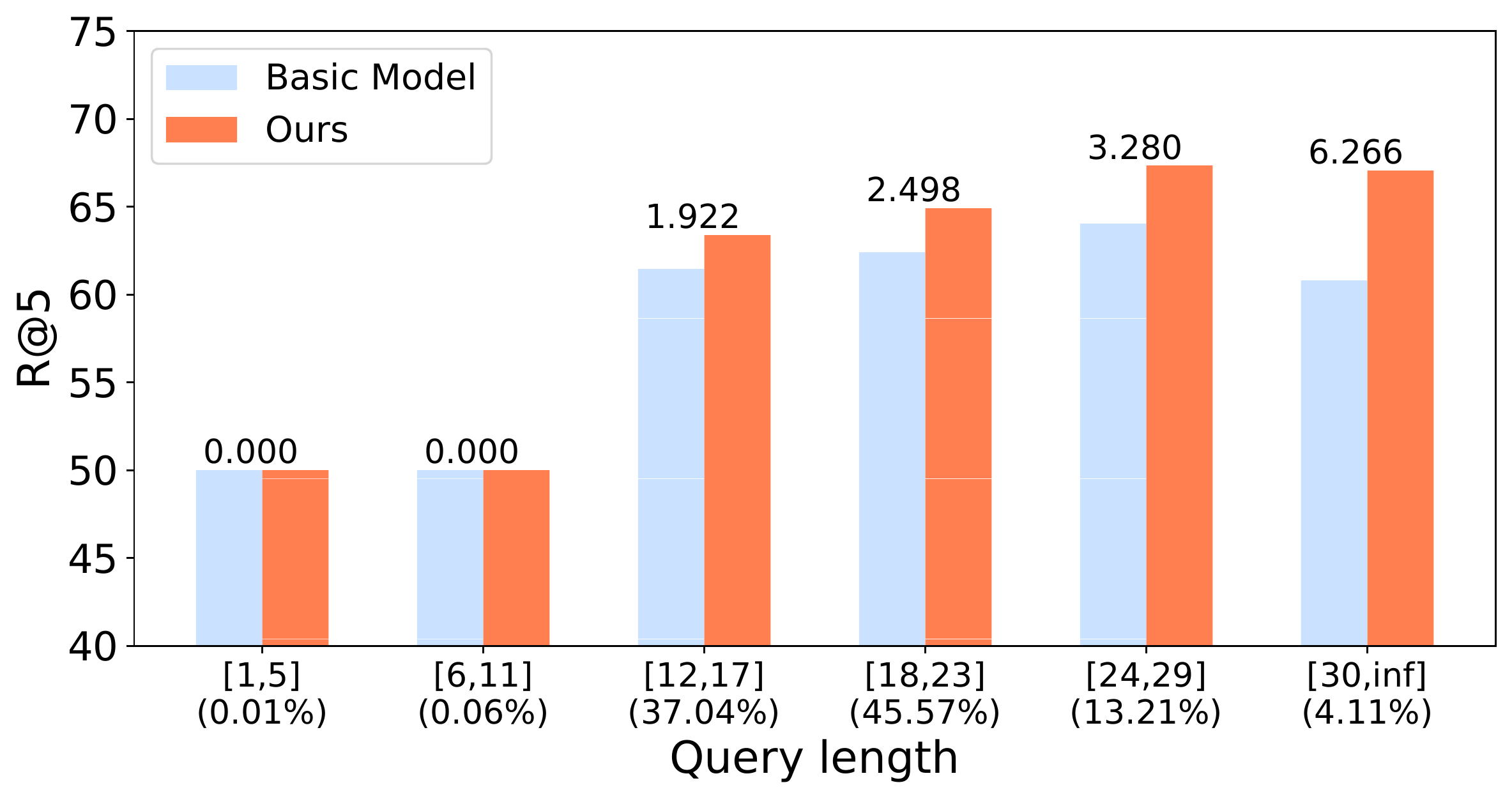}
\vspace{-3mm}
\caption{Performances of \textit{Basic Model} and  our NRCCR on queries of varied lengths. The number over the bins is the improvement of NRCCR over \textit{Basic Model}. $d\%$ represents the percentage of sentences with current length in VATEX.}\label{fig:length-group}
\end{figure}

\begin{table} [tb!]
\renewcommand{\arraystretch}{1.2}
\caption{\textbf{Ablation studies on VATEX to investigate the effectiveness of each model component.} 
}
\label{tab:ablation-vatex}
\centering 
\scalebox{0.9}{
\begin{tabular}{@{}l*{20}{r}c @{}}
\toprule
\multirow{2}{*}{\textbf{$L_{sim}$}} &\multirow{2}{*}{\textbf{$L_{feat}$}} & \multirow{2}{*}{\textbf{$L_{cyc}$}} & \multirow{2}{*}{\textbf{$L_{adv}$}}  & \multicolumn{2}{c}{\textbf{T2V}} && \multicolumn{2}{c}{\textbf{V2T}} & \multirow{2}{*}{\textbf{SumR}} \\
 \cmidrule{5-6}  \cmidrule{8-9} 
&&&& R@10 & mAP && R@10 & mAP & \\
\cmidrule{1-10}
&&&& 71.7&44.15 &&79.1 &31.43 & 352.2 \\
\checkmark&&&&72.8 &44.52 &&79.9 &31.10 & 354.6\\
&\checkmark&&& 73.8 &44.30 && 80.4 &31.50 & 356.4\\
\checkmark&\checkmark&&& 74.2&44.85 && 80.0 &32.74 &360.2\\
\checkmark&\checkmark&\checkmark &&74.4 &45.31 && 80.6 &31.83 & 361.4\\
\checkmark&\checkmark&\checkmark&\checkmark &\textbf{75.1}  &\textbf{45.64} &&\textbf{80.9} &\textbf{32.40} & \textbf{364.7} \\
\bottomrule
\end{tabular}
 }
\end{table}

\textbf{Noise robustness during test.} 
In this experiment, we group target-language sentences in the VATEX test set into 6 sets according to their lengths and report \black{the R@5 scores of text-to-video retrieval on each set.}
We assume that the length of the input sentence is directly proportional to the difficulty of translation, as translating long sentences is more likely to introduce noise. 
Figure ~\ref{fig:length-group} shows  the performance of \textit{Basic Model} and our model on queries with varied lengths.
Our model outperforms \textit{Basic Model} except on the first two sets, where the number of sentences is too small and the results lack statistical significance.
Besides, we observe that the gap between our model and \textit{Basic Model} widens when input sentences become longer, which further verifies the noise robustness of our model.

\subsection{Ablation Studies}
In this section, we investigate the contribution of each component in NRCCR.
The results on VATEX are summarized in Table \ref{tab:ablation-vatex}, where the first row reports the performance of \textit{Basic Model}.
Compared to \textit{Basic Model}, we empirically observe that self-distillation from the similarity-based view and feature-based view improves the performance by 2.4 and 4.2 on SumR metric, respectively.
In addition, performance is further improved by using both of them (the sumR increases from 352.2 to 360.2), which suggests the complementarity between these two views.
By additional minimizing the cycle semantic consistency loss $\mathcal{L}_{cyc}$ between source sentences and back-translated sentences, our model achieves an extra performance gain.
Finally, performing adversarial training is also beneficial to the overall performance, indicating the importance of learning language-agnostic representations.

\subsection{Analysis by t-SNE Visualization}\label{ssec:speed}

\begin{figure}[tb!]
\centering
\subfigure[Basic Model]{
\label{tsne-a}
\includegraphics[width=0.45\columnwidth]{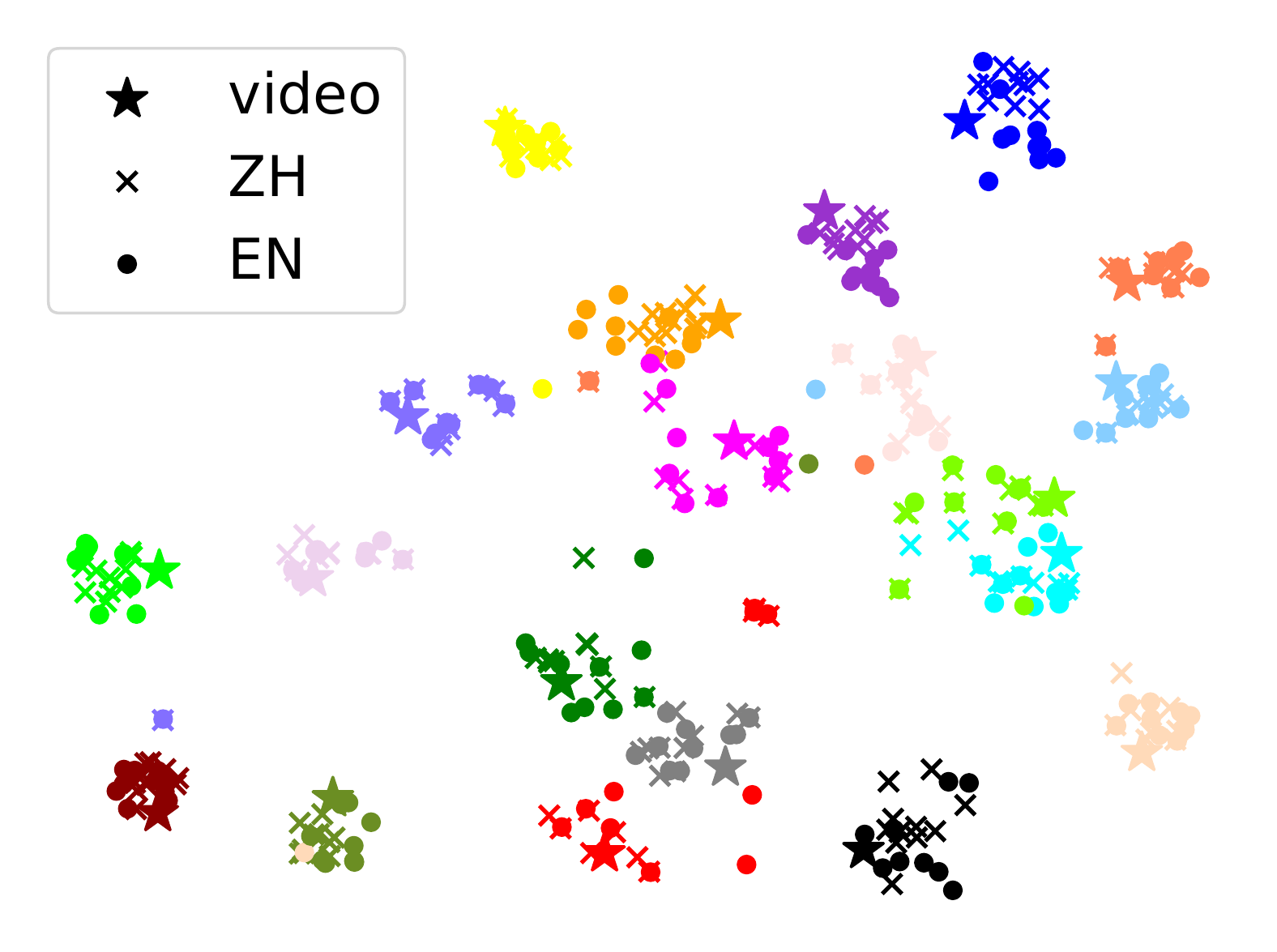}
}
\quad
\subfigure[Ours]{
\label{tsne-b}
\includegraphics[width=0.45\columnwidth]{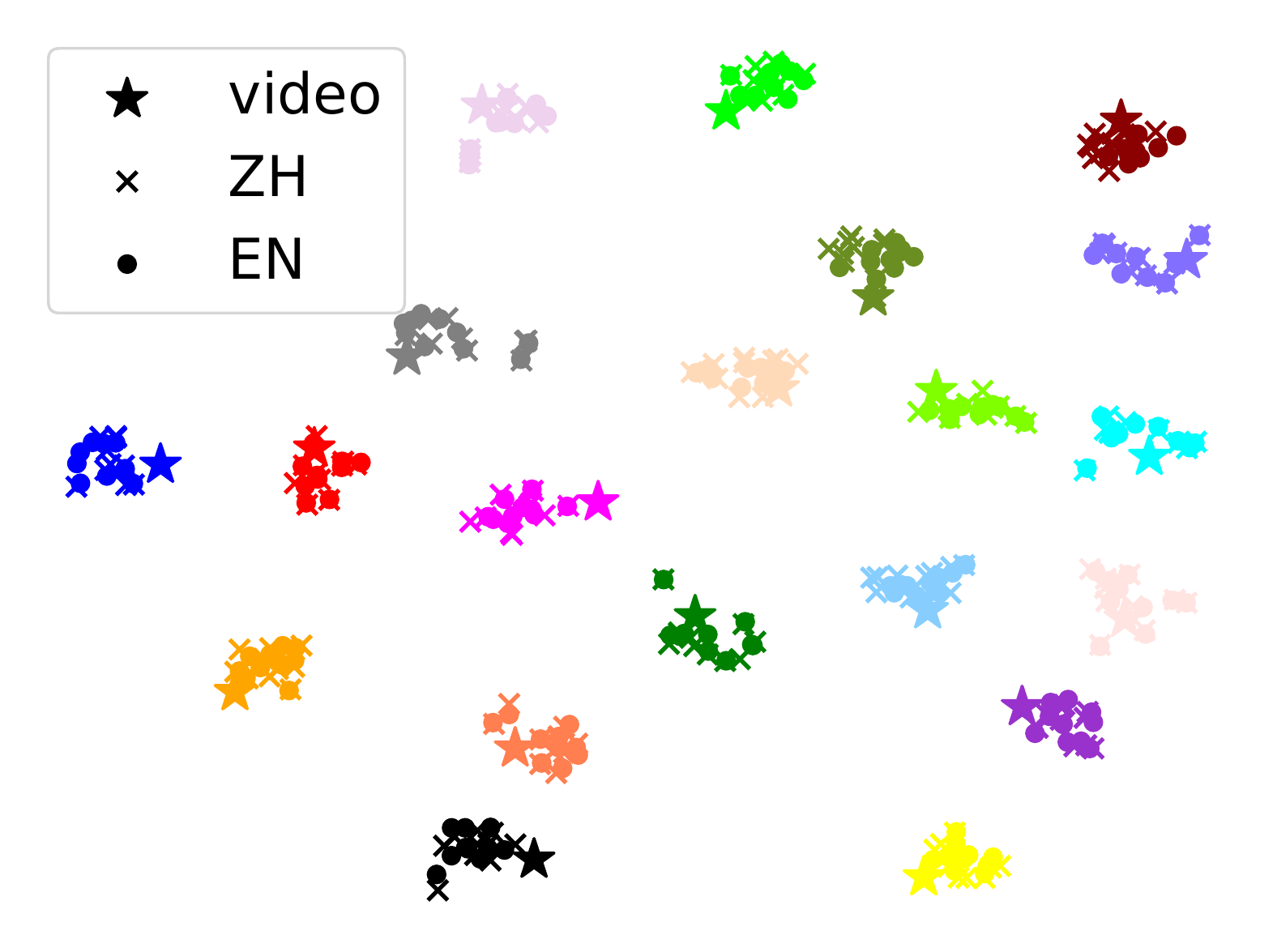}
}
\vspace{-3mm}
\caption{The t-SNE results produced by \textit{Basic Model} and NRCCR. 20 samples are randomly selected from VATEX and each of them is regarded as a class. 
Dots with the same color indicate  representations belonging to the same class. 
}\label{fig:tsne}
\end{figure}

To further analyze why our method yields better performance than \textit{Basic Model},
we randomly select 20 samples from the Chinese test set of VATEX and visualize their representations produced by \textit{Basic Model} (shown in Figure~\ref{tsne-a}) and NRCCR (shown in Figure~\ref{tsne-b}), respectively.
For each sample, it consists of a video, 10 Chinese sentences and 10 corresponding translated English sentences.
Compared with \textit{Basic Model}, the intra-class representations in NRCCR are relatively more compact,
which reveals that the language-agnostic representations have been learned and our model achieves a better cross-lingual cross-modal alignment.
It is worth noting that the English sentences obtained by MT are noisy, but the English representations are still closely clustered with Chinese representations in Figure~\ref{tsne-b}. 
This result to some extent demonstrates that our method is able to restore the original semantic information from noisy English sentences, and the reason might be that our method has been optimized by cycle semantic consistency learning.

\section{Summary and Conclusions} \label{sec:conc}
This paper proposes NRCCR, a noise-robust learning framework for CCR. We resort to MT to achieve cross-lingual transfer, and effectively alleviate the noise problem caused by MT in both training and testing stages via noise-robust learning. 
We conduct the experiments on two multilingual video-text datasets and a multilingual image-text dataset, outperforming previous CCR methods. 
The experimental results show that our method can effectively enhance the robustness of the model, and achieves significant improvement on alignment quality of cross-modal cross-lingual.
We hope our work could bring insights about cross-lingual transfer.
In future work, we would like to investigate the effectiveness of our model when transferring to multiple target languages simultaneously.

\section{Acknowledgments}
This work was supported by the National Key R\&D Program of China (No. 2018YFB1404102), NSFC (No. 61902347,
61976188), the Public Welfare Technology Research Project of Zhejiang Province (No. LGF21F020010), the Open Projects Program of the National Laboratory of Pattern Recognition, and the Fundamental Research Funds for the Provincial Universities of Zhejiang.

\bibliographystyle{ACM-Reference-Format}
\balance
\bibliography{mm2022}

\clearpage

\appendix

{
\begin{figure*}[htbp]
\centering\includegraphics[width=17.8cm]{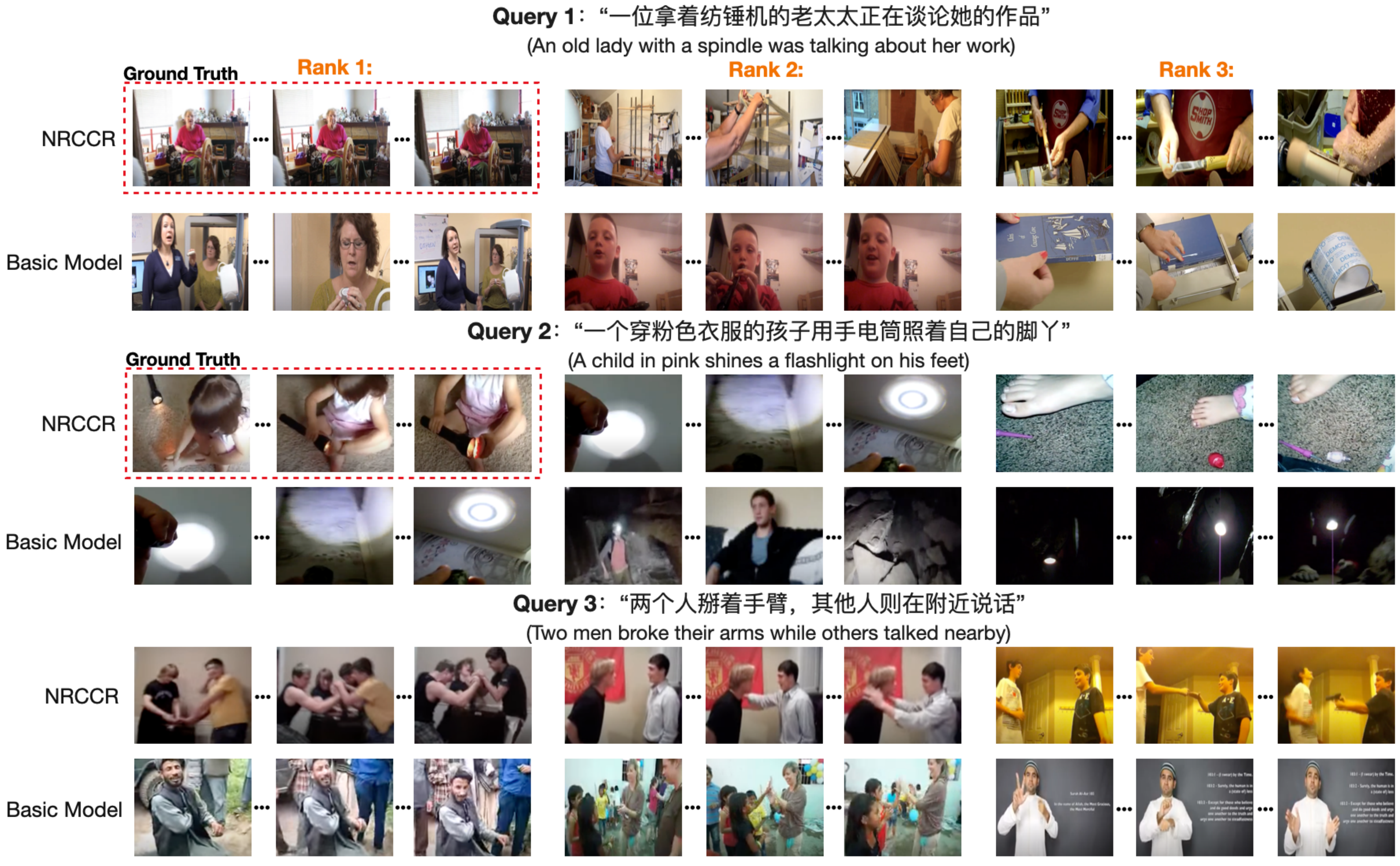}
\vspace{-8mm}
\caption{Examples of cross-lingual text-to-video retrieval on VATEX. We present top 3 videos retrieved for each query in the target language~(Chinese), given by \textit{Basic Model} and NRCCR respectively.
 }\label{fig:t2v}
\end{figure*}
\section{Cross-lingual Text-Video retrieval}

\subsection{Qualitative Experiment}
\label{sec: t2v_qualitative}

To qualitatively illustrate model behavior, Figure \ref{fig:t2v} lists three Chinese queries selected from VATEX and  top 3 videos retrieved by \textit{Basic Model} and NRCCR, respectively.
For  Query 1 and Query 2, NRCCR successfully ranks the ground truth as top 1 \black{and the top three retrieved videos are typically relevant to the given query to some extent}.
By contrast, \textit{Basic Model} does not  understand the query accurately and thus fails to answer these queries.
The reason might be that some noisy sentences generated by MT cannot correctly describe the corresponding video content, but \textit{Basic Model} is still forced to close the distances between them during training, which hurts the encoding of clean Chinese sentences badly.
The failure of \textit{Basic Model} in these cases reveals the importance of noise-robust learning for video-text retrieval in cross-lingual scenarios.
For Query 3, although both models fail to answer the query, compared with \textit{Basic Model}, the content of top 3 videos retrieved by NRCCR is still highly related to the query.
The bad case for NRCCR might suggest that  NRCCR can capture global semantic information contained in input sentences, but it still has limitations in fine-grained semantic understanding.

\begin{figure*}[tb!]
\centering\includegraphics[width=17.8cm]{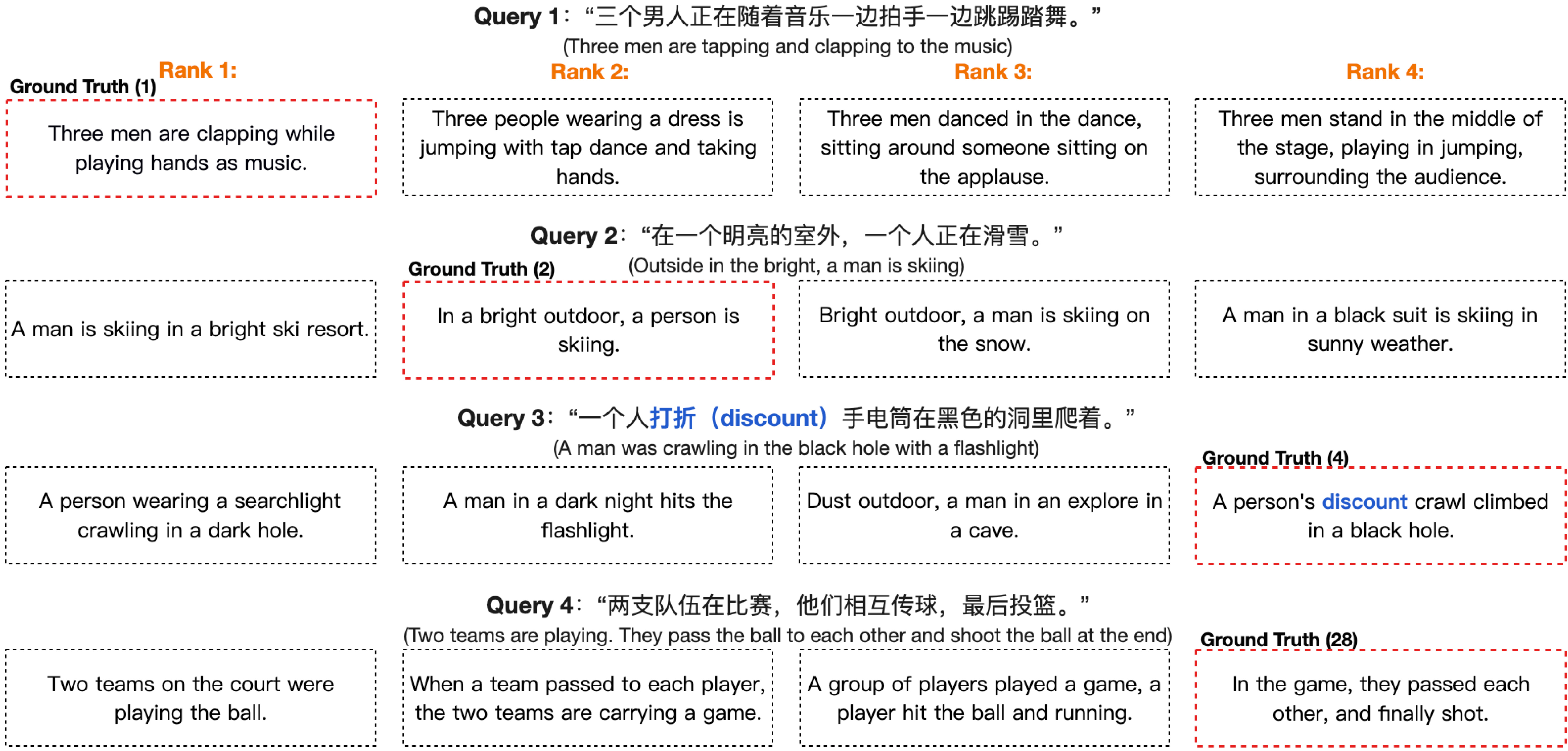}
\vspace{-7mm}
\caption{Qualitative results of cross-lingual text-to-text retrieval on VATEX. For each query, the top 3 ranked translated English sentences and the ground-truth sentences are shown. If the ground-truth sentence is among the top 3, the fourth sentence is included as well.}\label{fig:t2t}
\end{figure*}

\section{Cross-lingual Text-Text Retrieval}

\label{sec:t2t_retrieval}

\subsection{Quantitative Experiment}
\label{sec:t2t_quantitative}

To further verify the robustness of our model for text encoding, we conduct  quantitative analysis of cross-lingual text-to-text retrieval on the Chinese test set of VATEX. 
Specifically, following the setting in the main text, we first train NRCCR and \textit{Basic Model} on VATEX and translate the Chinese test set into English.
Given a Chinese sentence, if an English sentence is translated from it, we assume they are relevant, otherwise they are irrelevant.  
During evaluation, for each Chinese sentence, we retrieve its relevant English sentence according to the cosine similarity between sentence embeddings produced by NRCCR and \textit{Basic Model}
respectively. 
mAP is used as the retrieval metric and results are shown in Table \ref{tab:t2t}.
We observe that NRCCR significantly outperforms the \textit{Basic Model} by 19.44 on mAP, 
indicating that NRCCR achieves a robust cross-lingual alignment through noise-robust learning.

\begin{table} [htbp!]
\renewcommand{\arraystretch}{1.2}
\setlength{\abovecaptionskip}{0.2mm} 
\setlength{\belowcaptionskip}{-0.2cm}
\caption{\textbf{The quantitative comparison of cross-lingual text-to-text retrieval (ZH-to-EN). The performance is reported in percentage (\%)}. }
\label{tab:t2t}
\centering 
\scalebox{0.98}{
\begin{tabular}{@{}l*{12}{r}c @{}}
\toprule
Method &mAP \\
\cmidrule{1-2}
\textit{Basic Model} &61.84\\
NRCCR &81.28  \\
\bottomrule
\end{tabular}
 }
\end{table}

\subsection{Qualitative Experiment}
\label{sec:t2t_qualitative}

As shown in Figure \ref{fig:t2t}, we qualitatively analyze the results of the cross-lingual text-to-text retrieval.
For each Chinese query, only the English sentence machine translated from the query is regarded as the ground truth.
It is observed that top retrieved sentences in Figure \ref{fig:t2t} are semantically similar to the given query to some extent in Query 1 and Query 2.
In Query 3, the blue word is a typo produced during manual annotation, and the typo is machine translated as  "discount" in the ground truth without correction.
Given such a noisy query, however, the top retrieved sentences are all related to the correct meaning of the query.
This strongly confirms the robustness of our model, which could extract proper global semantic information from noisy sentences.
For Query 4, although our model fails to answer this query, 
the top retrieved sentences have a certain semantic relevance to the query.
This might suggest that our model pays more attention to coarse-grained global information than fine-grained information, as a result, it  has limitations in capturing fine-grained information (\eg objects and actions) in the input sentence, which has also been discussed in Section \ref{sec: t2v_qualitative}.

\section{Implementation Details}
\label{sec:implementation_details}
For video features, on MSR-VTT-CN, we use the frame-level features ResNet-152 \cite{he2016deep} and concatenate frame-level features ResNeXt-101 \cite{xie2017aggregated}\cite{mettes2020shuffled}  to obtain a combined 4,096-dimensional feature. On VATEX, we adopt the I3D \cite{carreira2017quo} video feature which is officially provided.
As for model structure, we employ a 1-layer Transformer with 8 heads and 1024 hidden units as the visual encoder. 
For text encoding, we use the pre-trained mBERT-base as the backbone and freeze layers below 7 as this setup empirically performs the best.
We let the heads of Transformer be 8, and only one block we use, which we find that a single block is able to achieves the expected performance.
For the optimization, we use PyTorch as our deep learning environment. The training batch size is 128.
An Adam optimizer with initial learning rate 1e-4 and adjustment schedule simialr to \cite{dong2022reading} is utilized. 
For  hyperparameters in Equation 12 (in the main text), we set $\lambda_1=0.4, \lambda_2=0.1, \lambda_3=0.5$ and $\lambda_4=1e-3$.
The hyperparameter $\alpha$ in the overall triplet loss is set to 0.6 and the weight $\beta$ used for inference is set to 0.8. Note that we will release our source code and data.

}









\end{document}